\documentclass{article}

\usepackage[preprint]{neurips_2022}

\usepackage[utf8]{inputenc} 
\usepackage[T1]{fontenc}    
\usepackage[colorlinks=true]{hyperref}       
\usepackage{url}            
\usepackage{booktabs}       
\usepackage{amsmath}
\usepackage{mathtools}
\usepackage{amsfonts}       
\usepackage{nicefrac}       
\usepackage{microtype}      
\usepackage{xcolor}         
\usepackage{graphicx}         
\usepackage{xspace}
\usepackage{wrapfig}

\newcommand{\name}{\textsc{NeuForm}\xspace}

\newcommand{\overfitted}[1]{\hat{#1}}
\newcommand{\frozen}[1]{\overline{#1}}
\newcommand{\edited}[1]{#1_E}
\newcommand{\original}[1]{#1_O}
\newcommand{\blended}[1]{\tilde #1}
\newcommand{\partset}{\mathcal{P}}
\newcommand{\ourname}{\textsc{NeuForm}\xspace}

\newcommand{\spaghetti}{\textsc{Spaghetti}\xspace}
\newcommand{\coalesce}{\textsc{Coalesce}\xspace}

\DeclareMathOperator*{\argmin}{arg\,min}

\title{\name: \\ Adaptive Overfitting for Neural Shape Editing}

\author{%
  Connor Zhizhen Lin \\
  Stanford University\\
  \texttt{connorzl@cs.stanford.edu}\\
  \And
  Niloy J.~Mitra\\
  Adobe Research and University College London\\
  \texttt{nimitra@adobe.com}\\
  \And
  Gordon Wetzstein\\
  Stanford University\\
  \texttt{gordonwz@stanford.edu}\\
  \And
  Leonidas Guibas\\
  Stanford University\\
  \texttt{guibas@cs.stanford.edu}\\
  \And
  Paul Guerrero \\
  Adobe Research \\
  \texttt{guerrero@adobe.com}\\
}

\begin{document}

\maketitle
\vspace*{-.2in}

\begin{abstract}
 Neural representations are popular for representing shapes, as they can be learned form sensor data and used for data cleanup, model completion, shape editing, and shape synthesis.  Current neural representations can be categorized as either overfitting to a single object instance, or representing a collection of objects. However, neither allows accurate editing of neural scene representations: on the one hand, methods that overfit objects achieve highly accurate reconstructions, but do not generalize to unseen object configurations and thus cannot support editing; on the other hand, methods that represent a family of objects with variations do generalize but produce only approximate reconstructions. We propose \name to combine the advantages of both overfitted and generalizable representations by adaptively using the one most appropriate for each shape region: the overfitted representation where reliable data is available, and the generalizable representation everywhere else. We achieve this with a carefully designed architecture and an approach that blends the network weights of the two representations, avoiding seams and other artifacts. We demonstrate edits that successfully reconfigure parts of human-designed shapes, such as chairs, tables, and lamps, while preserving semantic integrity and the accuracy of an overfitted shape representation. We compare with two state-of-the-art competitors and demonstrate clear improvements in terms of plausibility and fidelity of the resultant edits. 
\end{abstract}

\section{Introduction}

Neural formulations have emerged as an efficient and scalable representation of complex spatial signals, such as radiance fields, 3D occupancy fields, or signed distance functions. These representations are popular as they allow a uniform formulation that can support a range of applications including denoising, data completion, and editing. In the context of shapes, two main types of neural representations have emerged. Starting from an input description (e.g., point clouds, meshes, or distance/occupancy fields), current representations either overfit to a single shape or learn a model that generalizes over a collection of varying shapes. However, neither of the representations alone allows effective shape editing.

Overfitted models~\cite{davies2020overfit,takikawa2021neural,sitzmann_implicit_2020,mildenhall_nerf_2020,morreale2022neuralConvolutionalSurfaces,martel2021acorn,morreale2022neuralConvolutionalSurfaces} reproduce a single shape with high fidelity.
While this allows for operations like efficient rendering, surface-based optimization, and data compression, such a representation does not support shape editing or synthesis, since it does not generalize to novel shape configurations.

\begin{wrapfigure}[16]{l}{0.5\linewidth}
   \vspace{-10pt}
   \centering
   \includegraphics[width=\linewidth]{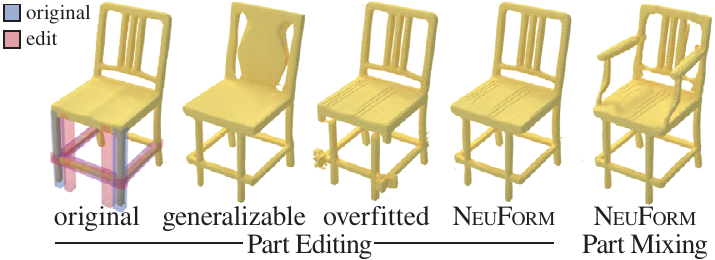}
   \caption{\textbf{Adaptive overfitting.} \name  enables detail preserving shape edits that generalize to new part configurations by combining advantages of a generalizable representation (e.g., generation of plausible joint geometry) and an overfitted representation (e.g., detail preservation on the backrest), and also allows mixing parts from different shapes.}
   \label{fig:teaser}
\end{wrapfigure}

In contrast, generalizable representations~\cite{park2019deepsdf,dualSDF20,occupancy_nets_2019,chen_cvpr19} are trained on a large collection of shapes and learn shape priors allowing 
the representation to adapt to previously unseen shape configurations. Thus, they can be used for shape editing and novel shape synthesis~\cite{hertz2022spaghetti,dualSDF20,deformSyncNet:2020,MoGuerreroEtAl:StructureNet:SiggraphAsia:2019,liu2018voxelgan,MoGuerreroEtAl:StructEdit:Arxiv:2019}. However, this comes at the cost of a lower-fidelity representation, as the network needs to represent a full dataset and its variations, instead of a single shape. Specifically, these models typically require `projecting' a shape into the learned latent space before editing it, where the idiosyncrasies of the starting model, in the form of local geometric details, are often lost (see Figure~\ref{fig:teaser}).

We propose a novel blended architecture, called \name, to combine the advantages of the two representations described above. Specifically, we retain distinctive properties of the input shape by relying on an overfitted model and switch to the generalizable model to complete parts where information is missing (e.g., near new joint locations or regions with holes). The main challenge is
to train an adaptive mixing network that
blends the information between the overfitted and generalizable models, without introducing artifacts such as undesirable seams or gaps.  The \name architecture allows this seamless sharing of information between the individual networks. Our main technical insights are that (i) it is possible to smoothly interpolate between two neural shape representations by blending between the weights of two networks sharing an architecture and a training history, and (ii) it is possible to do this blending between a generalizable network that works on a global view of the shape and an overfitted network that only has access to part of the shape by carefully pruning the information flow during overfitting. 



We evaluate \name on multiple applications: (i)~reconstruction (i.e., projecting a given input to an adaptive overfitted latent space); (ii)~part 
based shape editing; and (iii)~shape mixing (i.e., converting an arrangement of parts taken from different models into a coherent shape model). We compare with two state-of-the-art approaches~\cite{hertz2022spaghetti,yin2020coalesce} and demonstrate advantages, both quantitatively and qualitatively. Figure~\ref{fig:teaser} shows an example of a shape edit where we can see a clear advantage for \name over both purely generalizable and purely overfitted representations.

\section{Related Work}

\vspace{-5pt}
\paragraph{Single-Scene Neural Shape Representations}
Overfitting networks represent one specific shape via a single network by optimizing network weights. Such overfitted networks are useful for several applications including compression~\cite{davies2020overfit,takikawa2021neural}, adaptive network parameter allocation~\cite{martel2021acorn}, multiview reconstruction~\cite{sitzmann_implicit_2020,mildenhall_nerf_2020}, shape optimization~\cite{morreale2021neural}, or multi-resolution shape representation~\cite{morreale2022neuralConvolutionalSurfaces,yifan2021geometryconsistent,takikawa2021neural}. While such networks, by construction, accurately capture the original shapes, faithfully encoding their finer details, they can neither be used for editing shapes nor for creating new shapes by combining parts from multiple (source) shapes.

\vspace{-5pt}
\paragraph{Multi-Scene Neural Shape Representations}

Neural networks have been used to approximate implicit models, as an example of complex spatial functions, to represent  shapes as  volumetric signed distance fields~\cite{park2019deepsdf,chen_cvpr19} or occupancy values~\cite{occupancy_nets_2019}. Such network learning has been further regularized by geometric constraints like the Eikonal equation ~\cite{gropp_implicit_2020,atzmon_sal_2020,atzmon_sal_2019} or using an intermediate meta-network for faster convergence~\cite{littwin_deep_2019}.  Other approaches model shapes using their 2D parameterizations~\cite{groueix2018papier,yang_foldingnet_2018}. Improved versions of such methods optimize for low-distortion atlases~\cite{bednarik_shape_2019}, learn task-specific geometry of 2D domain~\cite{deprelle_learning_2019,Sinha2016DeepL3}, or force the surface to agree with an implicit function~\cite{Poursaeed20a}. Most of these methods encode shape collections in a lower-dimensional latent space, as a proxy for the underlying shape space, and support shape editing and generative modeling. For example, sampling from and optimizing in the (restricted) latent spaces can produce voxel grids~\cite{liu2018voxelgan,GirdharFRG16,BrockLRW16,dai2017complete}, point clouds~\cite{achlioptas2018latent_pc,Su2017PointGen},   meshes~\cite{dai2019scan2mesh}, or collections of deformable primitives~\cite{genova2019}.
Others~\cite{dualSDF20,MoGuerreroEtAl:StructureNet:SiggraphAsia:2019,deformSyncNet:2020} use a two level representations with a primitive-based coarse structure capturing the part arrangement, and a detailing network that adds high-resolution part level geometric details.  While these methods do generalize across shapes, and can be used for editing~\cite{dualSDF20,deformSyncNet:2020,hertz2022spaghetti,yin2020coalesce}, the source models often lose their finer details during the projection to the underlying latent space and subsequent editing process. In Section~\ref{sec:results}, we compare against two of the most relevant methods: \coalesce~\cite{yin2020coalesce}, which focuses on part-based modeling and synthesizing part connections (i.e., joints), and \spaghetti~\cite{hertz2022spaghetti}, which focuses on inter-part relations towards shape editing and mixing. Our method, \ourname, generates higher quality joints than the former while preserving more (original) surface detail than the latter.

\section{Method}

\begin{wrapfigure}[26]{r}{0.54\linewidth}
    \vspace{-15pt}
    \centering
    \includegraphics[width=\linewidth]{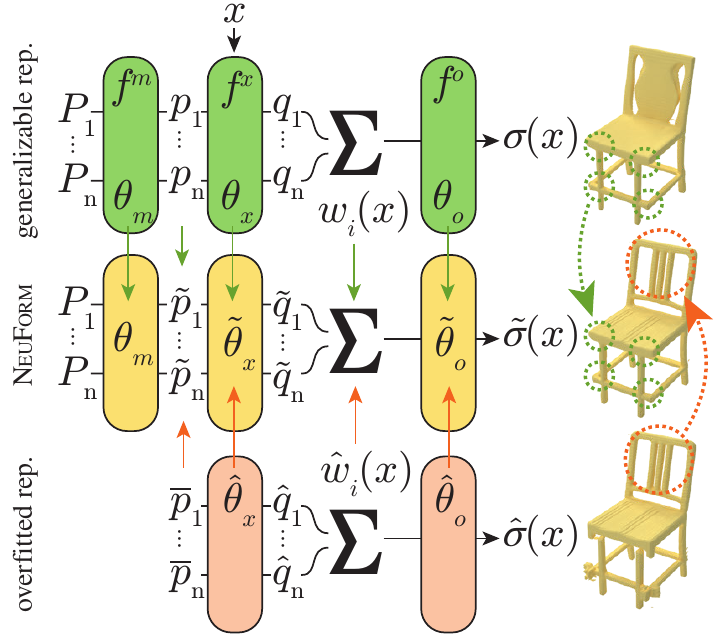}
    \caption{\textbf{Architecture overview.} \name blends between a generalizable neural shape representation (green) and an overfitted neural shape representation (red) by interpolating their network weights and some feature layers. This combines the benefits of detail preservation from the overfitted representation and editability from the generalizable representation.}
    \label{fig:architecture}
\end{wrapfigure}

Given a manifold and watertight 3D shape $S$ with known part annotations, our goal is to edit the parts of $S$ without introducing objectionable artifacts or losing geometric detail. The shape can be given as a mesh, signed distance function, or occupancy function, and the part annotations are specified as a set of oriented cuboid bounding boxes $\{C_1, \dots, C_n\}$, where $n$ is the number of parts of $S$. During editing, parts may be re-arranged via scaling and translation, and/or mixed across multiple shapes. To avoid artifacts in the edited shape, some regions of the shape geometry, such as the joints between individual shape parts, need to be adjusted to adapt to the new part configuration. To enable part-based editing without losing geometric detail, we construct two neural representations of shape $S$: a generalizable shape representation and an overfitted shape representation.

The \emph{generalizable shape representation} is a part-aware neural shape representation trained to represent a large shape space. This parameterization can generalize to previously unseen part configurations, including the edited configuration of shape $S$, but can only provide a low-fidelity reconstruction of $S$. 

The \emph{overfitted shape representation} is a neural shape representation overfitted to a single shape $S$. It represents the input shape geometry in great detail, but does not generalize to unseen part configurations, such as edited configurations of $S$. 

We combine these representations by blending between them, as explained in Section~\ref{sec:adaptive}. In regions where reliable data is available for overfitting, such as regions unaffected by edits, we use the overfitted shape representation. In regions where geometry should be adjusted, e.g. joint regions between parts, we leverage the generalizable representation. Both representations share the same architecture and we blend between them by directly interpolating their network parameters, which requires careful design of both the architecture and overfitting setup. We call this approach \emph{adaptive overfitting}. 

\paragraph{Generalizable Shape Representation}
\label{sec:generalizable}
\paragraph{Shape parameters.} In the generalizable representation, a shape $S$ is represented as a set of part parameters $\partset := \{P_1, \dots, P_n$\}. The parameters of a part $P_i := (C_i, g_i)$ consist of a cuboid bounding box $C_i:=(v_i, e_i, o_i)$, where $v_i, e_i, o_i$ are the centroid position, size, and orientation of the cuboid, respectively, and a latent vector $g_i$ defining the part's geometry in the local coordinate frame of the cuboid. We obtain $g_i$ from $S$ by encoding $m$ surface and volume points $r^i_1, \dots, r^i_m$ sampled from part $P_i$ with a PointNet~\cite{pointnet} encoder as $g_i := h_{\psi}(r^i_1, \dots, r^i_m)$, although other options to obtain $g_i$ such as an auto-decoder setup with inference time optimization are also possible.

\paragraph{Generalizable occupancy function.} Given the part parameters $P$, a neural network $f_\theta$ models the occupancy field $\sigma_S$ of shape $S$ at any query location $x$ as, 
\begin{equation}
    \sigma_S(x) \approx \sigma_\partset(x) := f_\theta(x|\partset). 
\end{equation}
%
%
The architecture of $f$ is illustrated in Figure~\ref{fig:architecture}. This is similar to the formulation proposed in \spaghetti~\cite{hertz2022spaghetti}, but  with changes that are necessary for adaptive overfitting. The network is composed of three parts: A part mixing network $f^m_{\theta_m}$ to exchange information between per-part latent vectors; a part query network $f^x_{\theta_x}$ to query each part at the query point $x$; and a global occupancy network $f^o_{\theta_o}$ aggregating the results of the per-part queries and output the occupancy at $x$.


\textit{(i) Part mixing network.}
The mixing network $f^m$ first converts parameters $P_i$ into per-part latent vectors, and then exchanges information between parts using a self-attention layer:
\begin{equation}
  p^\partset_i := f^m_{\theta_m}(P_i | \partset). 
\end{equation}

\textit{(ii) Part query network.}
The part query network $f^x$ queries each part $p^\partset_i$ at the local query point locations using cross-attention from each local query point to all per-part latent vectors $p^\partset_i$:
\begin{equation}
    q^\partset_i(x) := f^x_{\theta_x}(T^{-1}_{C_i}(x)\ |\ p^\partset_1+b_0, \dots, p^\partset_i+b_1, \dots, p^\partset_n+b_0),
\end{equation}
where
$T^{-1}_{C_i}$ denotes the transformation to the local coordinate frame of $C_i$. 
Like $f^m$, $f^x$ is run once per part. For a given part $i$, we augment the input latent vectors $p^\partset_*$
by adding a learned indicator feature that equals $b_1$ for the current part $i$ and $b_0$ for all other parts, giving the network knowledge of which parts it is currently processing. 
The resulting latent vector $q^\partset_i$ encodes the local geometry region of part $i$ that is relevant to the query point $x$.


\textit{(iii) Global occupancy network.}
Finally, we aggregate the per-part latent vectors $q^\partset_i$ into a global latent vector using a weighted sum and the global occupancy network $f^o$ computes the occupancy at the query location $x$:
\begin{equation}
    \sigma_\partset(x) := f^o_{\theta_o}\big(\sum_i w^\partset_i(x)\ q^\partset_i(x)\big),
\end{equation}
where the weights $w^\partset_i = \kappa\big(\max(0,d^s_i(x, C_i))\big)$ are based on the signed distance $d_i^s$ from query point to cuboid $C_i$.
We choose the triweight kernel for $\kappa$ as it combines a finite support with a smooth falloff: $\kappa(a_i) = (1-(\frac{a_i}{\rho})^2)^3$, where $\rho$ is the radius of the kernel and $a_i = \min(\max(d^s_i, 0), \rho)$ is the bounded distance to cuboid $C_i$. Essentially, $\rho$ defines the extent of joint regions and $\kappa$ provides a smooth fall-off to 0 as $a_i$ approaches $\rho$. We set $\rho=0.35$ in all our experiments.

\paragraph{Training setup.} We jointly train the part encoder $h_{\psi}$ and the occupancy network $f_\theta$  on a large dataset of shapes $\mathcal{S}$ using a binary cross-entropy loss between the predicted occupancy $\sigma_\partset(x)$ and the ground truth occupancy $\sigma_S(x)$. More details are given in the supplementary material.

\paragraph{Shape editing.} Due to training on a large dataset, the generalizable shape representation captures a large space of part configurations. Shape edits can be performed by modifying the parameters of one or multiple cuboids, such as the position $v_i$ or scale $e_i$, to obtain the modified part set $\edited{\partset}$ 
and infer a modified occupancy as $\sigma_{\edited{\partset}}(x) := f_\theta(x|\edited{\partset})$.

\paragraph{Overfitted Shape Representation}
\label{sec:overfitted}

\paragraph{Overfitted occupancy function.}
The goal of the overfitted representation is to accurately capture the geometric detail of individual parts of a single shape. We use an overfitted occupancy function $\hat{f}$ with the same architecture as in the generalizable representation to facilitate blending between the two, as described in the next section.
Naively overfitting this occupancy function to a shape $S$ would result in artifacts when reconstructing an edited shape $\edited{S}$, since the overfitted occupancy function does not generalize to unseen part configurations. Instead, we carefully sever the information flow between parts during overfitting such that querying the overfitted occupancy function does not use information about the full edited part configuration.
We employ a two-part strategy: 
(i) We freeze the part latent vectors $p^\partset_i$ before overfitting and only update the query network $f^x$ and the occupancy network $f^o$:
\begin{gather}
    \overfitted{\sigma}_\partset(x) = \overfitted{f}_{\overfitted{\theta},\frozen{\partset}}(x|\partset) = f^o_{\overfitted{\theta}_o}\big(\sum_i \overfitted{w}^\partset_i(x)\  \overfitted{q}^{\partset,\frozen{\partset}}_i(x)\big), \\
    \text{with }\hspace{3pt} \overfitted{q}^{\partset,\frozen{\partset}}_i (x)= f^x_{\overfitted{\theta}_x}(T^{-1}_{C_i}(x)\ |\ \frozen{p}^{\frozen{\partset}}_1+b_0, \dots, \frozen{p}^{\frozen{\partset}}_i+b_1, \dots, \frozen{p}^{\frozen{\partset}}_n+b_0), \nonumber
\end{gather}
where $\overfitted{\sigma}$ is the occupancy predicted by the overfitted network, ${\overfitted{\theta}_o}$, ${\overfitted{\theta}_x}$ are the overfitted parameters of the query and occupancy networks, and $\frozen{p}^{\frozen{\partset}}$ denotes part latent vectors that were frozen to the part set ${\frozen{\partset}}$.
%
(ii)~We change the weights $\overfitted{w}^\partset_i$ to only select the single part latent vector $q^\partset$ that is closest to the query point $x$: $\overfitted{w}^\partset_i(x) = \mathbf{1}_{\{i\}}(\argmin_i d^s_i(x, C_i))$, where $\mathbf{1}$ is the indicator function.
These two changes effectively make the occupancy $\overfitted{\sigma}_\partset(x)$ at each query point dependent on only the single closest part, preventing the overfitted occupancy function from being exposed to an unseen part configuration.

\paragraph{Training setup.}
We start with a trained generalizable network $f_\theta$ and a part set $\frozen{P}$ we would like to overfit to. We freeze the part latent vectors $\frozen{p}^{\frozen{\partset}}_i = f^m_{\theta_m}(P_i|\frozen{\partset})$ to the values computed by the generalizable network, and then proceed to overfit both $f^x_{\theta_x}$ and $f^o_{\theta_o}$ to the partset $\frozen{\partset}$, giving us the overfitted network $\overfitted{f}_{\overfitted{\theta},\frozen{\partset}}$.
During overfitting, we gradually blend between the original weights $w_i$ at the first epoch to the updated weights $\overfitted{w}_i$ at the last epoch.

\paragraph{Shape editing.}
Similar to the generalizable representation, edits of the overfitted representation can be performed by modifying cuboid parameters to obtain a modified part set $\edited{\partset}$, 
and a modified occupancy $\overfitted{\sigma}_{\edited{\partset}}(x) = f_{\overfitted{\theta},\frozen{\partset}}(x|\edited{\partset})$. As a result of our strategy to decouple parts from each other, a transformation $T_i$ of a cuboid $C_i$ is directly applied to the occupancy of the corresponding part: $\overfitted{\sigma}_{\edited{\partset}}(x) = \overfitted{\sigma}_{\partset}(T^{-1}_i(x))$ for all $x$ that are closer to cuboid $i$ than to any other cuboid. This accurately preserves geometric detail after an edit, but results in discontinuities at the boundaries between edited parts, as shown in Figure~\ref{fig:teaser}. 







\paragraph{Adaptive Overfitting}
\label{sec:adaptive}
Our goal is to use the overfitted representation in areas where the overfitted occupancy is reliable, and the generalizable representation everywhere else. For shape edits that transform cuboid parameters, the overfitted occupancy in any local region undergoes the same transformation as the nearest cuboid. For human-made shapes such as chairs and tables, this behaviour is desirable in regions that are either close to only one cuboid, or close to only unedited cuboids. In other regions
(near joints between two or more cuboids, or where at least one cuboid has been edited),
the occupancy may need to undergo more complex transformations to reflect the new part configuration.

Given a set of parts $\original{\partset}$ and an edited version of the parts $\edited{\partset}$, we formalize the intuition described above as a scalar blending field $\lambda(x)$ defining a blending factor in $[0,1]$ between the generalizable and the overfitted representation at each query point $x$: 
\begin{equation}
\label{eq:blending_field}
    \lambda(x) := \kappa\big(\min_{C\ \in\ (\mathcal{C}^{\original{\partset}}\ \cup\ \mathcal{C}^{\edited{\partset}} / C^{\edited{\partset}}_\text{min})} d^s_i(x, C)\big), 
\end{equation}
where $\mathcal{C}^{\original{\partset}}_E$ and $\mathcal{C}^{\edited{\partset}}_E$ are the subsets of cuboids in the original and edited shape, respectively, that have been changed in $\edited{\partset}$. $C^{\edited{\partset}}_\text{min}$ is the cuboid in $\edited{\partset}$ closest to $x$. The kernel $\kappa$ is the same triweight kernel defined in Section~\ref{sec:generalizable} for part aggregation in the global occupancy network.

Given a blending factor $\lambda(x)$, we finally fuse the two representations by blending between the parameters, weights, and features of the networks:
\begin{gather}
    \blended{\sigma}_{\partset}(x) := f^o_{\blended{\theta}_o}\big(\sum_i \blended{w}^\partset_i(x)\  \blended{q}^{\partset,\frozen{\partset}}_i(x)\big), \\
    \text{with }\hspace{3pt} \blended{q}^{\partset,\frozen{\partset}}_i (x)= f^x_{\blended{\theta}_x}(T^{-1}_{C_i}(x)\ |\ \blended{p}^{\partset,\frozen{\partset}}_1+b_0, \dots, \blended{p}^{\partset,\frozen{\partset}}_i+b_1, \dots, \blended{p}^{\partset,\frozen{\partset}}_n+b_0), \nonumber
\end{gather}
where $\blended{\theta_o}$, $\blended{\theta_x}$, $\blended{w}^\partset_i(x)$, and $\blended{p}^{\partset,\frozen{\partset}}$ are linearly interpolated between the overfitted and generalizable representation using the blending factor $\lambda(x)$:
\begin{align}
\blended{\theta_*} &= (1-\lambda(x))\ \overfitted{\theta}_* + \lambda(x)\ \theta_*, \\
\blended{w}^\partset_i(x) &= (1-\lambda(x))\ \overfitted{w}^\partset_i(x) + \lambda(x)\ w^\partset_i(x), \\
\blended{p}^{\partset,\frozen{\partset}}_i &= (1-\lambda(x))\ \frozen{p}^{\frozen{\partset}}_i + \lambda(x)\ p_i^\partset(x).
\end{align}
When editing a shape, we typically overfit to the original configuration of the parts, in that case, we set $\frozen{\partset} = \original{\partset}$ and $\partset = \edited{\partset}$.

\begin{figure}[t]
    \centering
    \includegraphics[width=\linewidth]{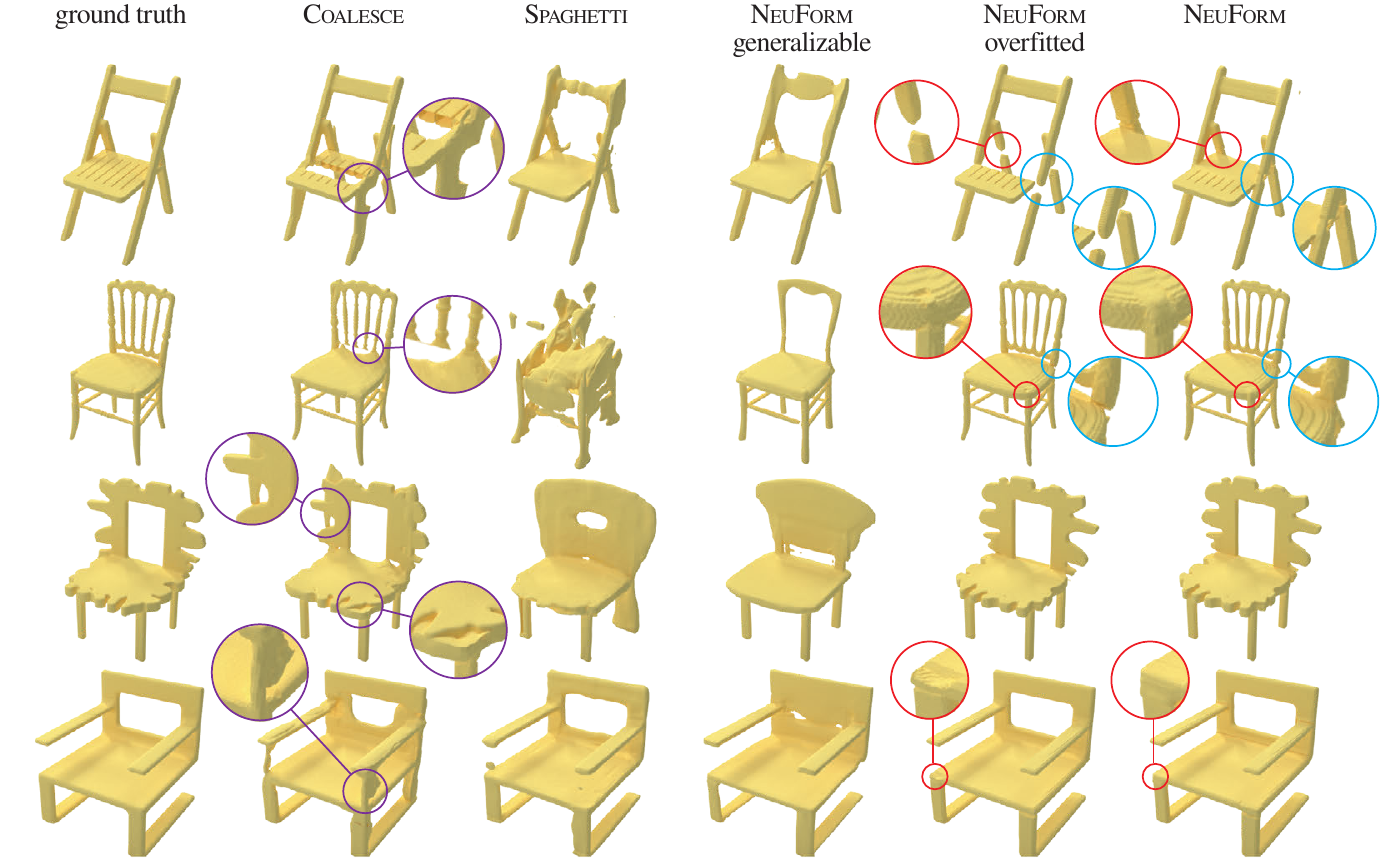}
    \caption{
    \textbf{Shape reconstruction.} 
    Comparing reconstructions of PartNet~\cite{Mo_2019_CVPR_partnet} \texttt{chairs}. We show reconstructions of four shapes. \coalesce and the overfitted representation preserve geometric detail, but have more artifacts near joints. \spaghetti and the generalizable representation perform better near joints but lose geometric detail. \name combines the best of both worlds.}
    \label{fig:comparison_chair_recons}
\end{figure}

\section{Results}
\label{sec:results}
We evaluate \name on three tasks: shape reconstruction,  
shape editing, 
and shape part mixing.  

\paragraph{Dataset.}
We use the PartNet~\cite{Mo_2019_CVPR_partnet} dataset for our experiments. PartNet is a dataset of human-made shapes in $24$ common categories, including furniture and typical household items. Each shape is annotated with hierarchical part segmentation. We experiment on the \texttt{chair}, \texttt{lamp}, and \texttt{table} categories and select hierarchy levels that result in an average of roughly $8$, $4$, and $8$ parts for \texttt{chairs}, \texttt{lamps}, and \texttt{tables}, respectively. Cuboids are computed as oriented bounding boxes of the segmented parts using Trimesh~\cite{trimesh}. We train the generalizable model on each shape category separately and choose a training/test split of $6000/1800$, $2100/400$, and $3500/500$ for \texttt{chairs}, \texttt{lamps}, and \texttt{tables}, respectively. All shapes are centered and the largest bounding box side is scaled to $2$.

\paragraph{Training details.}
We train the generalizable model for $1000$ epochs using the Adam~\cite{kingma2015adam} optimizer with a learning rate of $1e-4$ and an exponential learning rate decay of 0.994 per epoch. 
In each epoch, we train on 4096 query points per shape with a batchsize of 1 shape. We sample $12.5\%$ of the points uniformly in the $[-1,1]$ cube and $87.5\%$ of the points around the surface with a Guassian offset ($\mathcal{N}(0,0.05)$).
The overfitted model is trained for $100$ epochs on a single shape using the same training setup. Training the generalizable model takes roughly 33 hours on a TitanXp GPU and training the overfitted model takes roughly $25$ minutes on a single V100 GPU.

\paragraph{Baselines and ablations.}
We compare our results to \spaghetti~\cite{hertz2022spaghetti} as the state-of-the-art generalizable representation sharing a similar architecture to our generalizable representation, and \coalesce~\cite{yin2020coalesce}, a state-of-the-art method generating the joint geometry between parts given (potentially re-arranged) part meshes. Additionally, we compare with two ablations of our method: using only the generalizable representation and using only the overfitted representation.

\begin{figure}[t!]
    \centering
    \includegraphics[width=\linewidth]{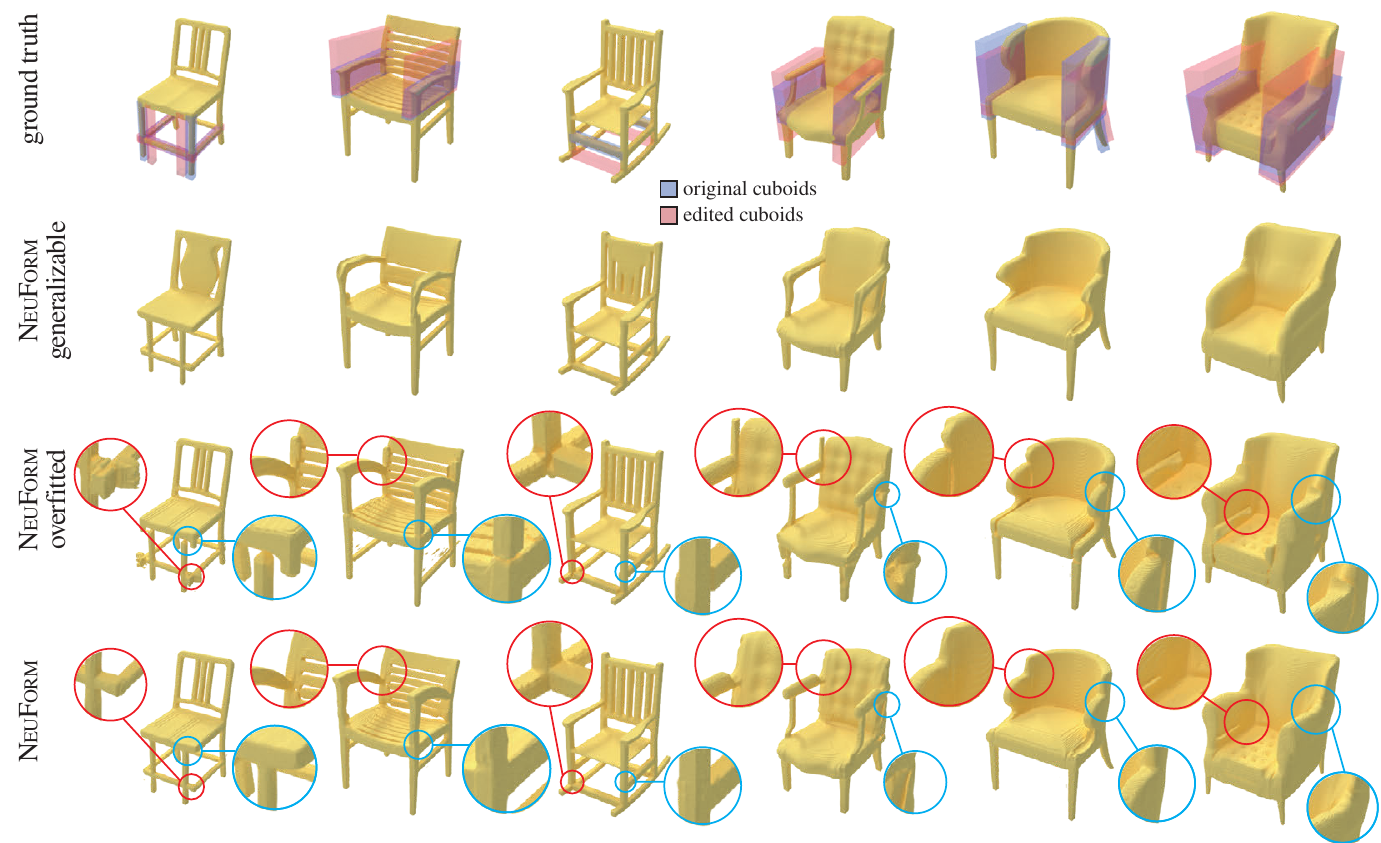}
    \caption{
    \textbf{Shape editing.} 
    Comparing edits on PartNet \texttt{chairs} when using only the generalizable or only the overfitted representations.
    We show edits on shapes with different coarse structure and fine scale details. The generalizable representation has plausible joint areas, but lacks geometric detail;  the overfitted representation preserves detail, but has artifacts near joints (see zoom-ins). \name combines the two representations to both preserve geometric detail and generate plausible joints.}
    \label{fig:comparison_chair_edits}
\end{figure}

\paragraph{Metrics.}
As quantitative metrics, we follow prior work in using the Chamfer Distance (CD) and Earth Mover's Distance (EMD) between points sampled on generated shape surface and points sampled on ground truth shape surfaces. For CD, we sample $30k$ and $10k$ points uniformly on the shape surfaces away from and near joint regions, respectively. We sample $1024$ points away from and near joint regions for EMD. As a volumetric measure, we evaluate the signed distance field (SDF) at $25k$ points away from joint regions and $5k$ points near joint regions per shape, with the same distribution as the query points, and report the absolute difference between the values of the generated and ground truth shapes. Since our tasks focus on the joints between shape parts, we separately report these metrics on joint regions ($\lambda(x) < 0.5$; see Eq.~\ref{eq:blending_field}), non-joint regions, and an unweighted average of the two.

\paragraph{(i) Shape Reconstruction.}
\label{sec:shape_reconstruction}

\begin{table}[b!]
  \small
  \centering
  \caption{
  Comparing shape reconstruction performance. We compare our results to all baselines and ablations. The Chamfer Distance is multiplied by $10^2$. \spaghetti and our generalizable representation perform well in joint regions, while \coalesce and the overfitted representation perform better in non-joint regions. The adaptive overfitting performed by \name achieves good performance in both regions, resulting overall in a significant improvement over both \spaghetti and \coalesce. As one would expect, the overfitted representation performs perticularly well on the reconstruction task, but its performance on joint regions drops significantly in shape editing tasks, as we demonstrate qualitatively in the following sections.
  }
  \setlength{\tabcolsep}{2pt}
  \begin{tabular}{r @{\hskip 20pt} ccc @{\hskip 20pt} ccc @{\hskip 20pt} ccc}
    \toprule
    & \multicolumn{3}{c}{Joint regions} & \multicolumn{3}{c}{Non-joint regions} & \multicolumn{3}{c}{All regions}\\
    \midrule
    & CD$\downarrow$ & EMD$\downarrow$ & SDF$\downarrow$ & CD$\downarrow$ & EMD$\downarrow$ & SDF$\downarrow$ &  CD$\downarrow$ & EMD$\downarrow$ & SDF$\downarrow$ \\ 
    \midrule
    \spaghetti~\cite{hertz2022spaghetti} & 0.337 & \textbf{65.54} & \textbf{1.343} & 1.381 & 176.27 & 3.758 & 0.859 & 120.96 & 2.570 \\
    \coalesce~\cite{yin2020coalesce} & 0.738 & 97.51 & 2.440 & \textbf{0.154} & 130.20 & 2.918 & 0.446 & 113.86 & 2.679 \\
    \midrule
    \ourname generalizable & 0.390 & 84.27 & 2.109 & 0.523 & 117.81 & 5.208 & 0.457 & 101.04 & 3.659 \\
    \ourname overfitted & 0.318 & 78.54 & 2.198 & 0.157 & \textbf{80.45} & 2.644 & \textbf{0.238} & \textbf{79.50} & 2.471 \\
    \ourname & \textbf{0.253} & 78.05 & 1.814 & 0.334 & 88.53 & \textbf{2.538} & 0.293 & 83.29 & \textbf{2.176} \\
    \bottomrule
  \end{tabular}
  \label{tab:comparison_chair_recons} 
\end{table}

First, we evaluate the reconstruction performance of \name compared to the baselines and ablations on $64$ shapes selected randomly from the test set. \coalesce does not support fine-grained parts, thus, for a fair comparison, we restrict our joint areas to those defined by \coalesce in this experiment. Our overfitted model is trained without ground truth for any of the joint areas.

\begin{figure}[t!]
    \centering
    \includegraphics[width=\linewidth]{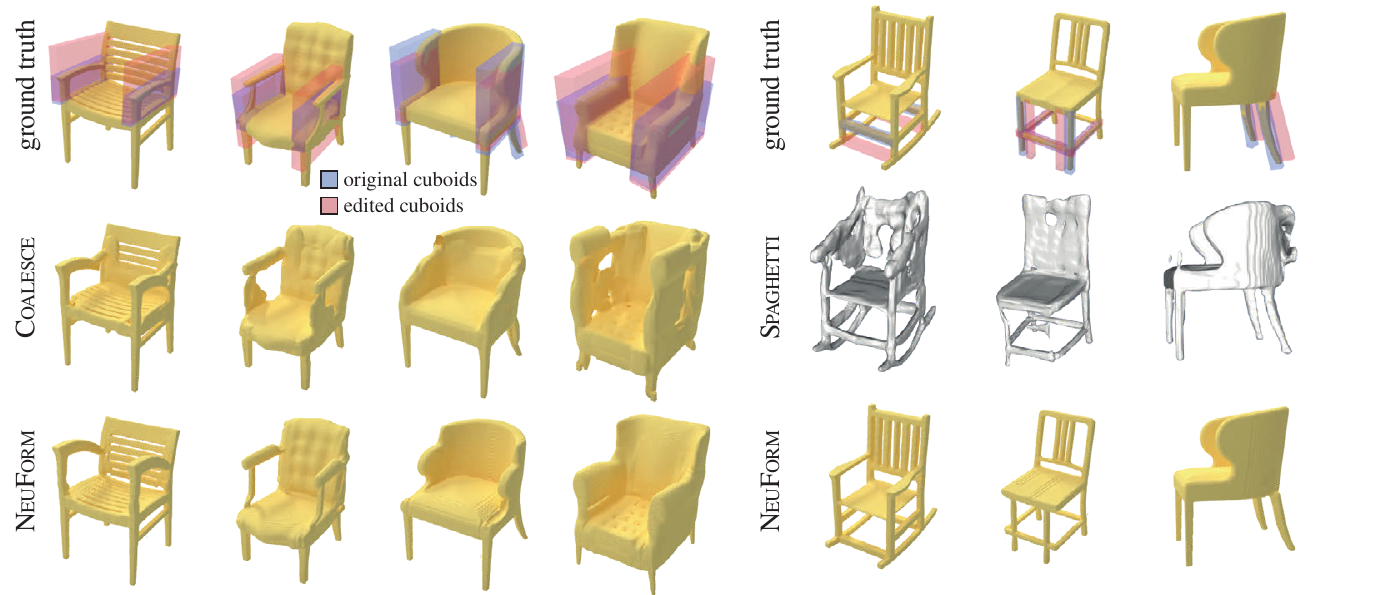}
    \caption{Comparing edits on PartNet \texttt{chairs} to \coalesce~\cite{yin2020coalesce} and \spaghetti~\cite{hertz2022spaghetti}. We show two different sets of edits because \coalesce does not support edits of more fine-grained parts like bars, while \spaghetti does not currently support part scaling in their released code. \coalesce struggles with more extended joint areas and \spaghetti's result is significantly noisier after an edit. Here we show screenshots from \spaghetti's editing UI (hence the different color). Blending between the generalizable and overfitted representations using \ourname gives us more plausible edit results, with cleaner joints and detailed part geometry.}
    \label{fig:comparison_chair_edits_coalesce_spaghetti}
\end{figure}

\begin{figure}[b!]
    \centering
    \includegraphics[width=\linewidth]{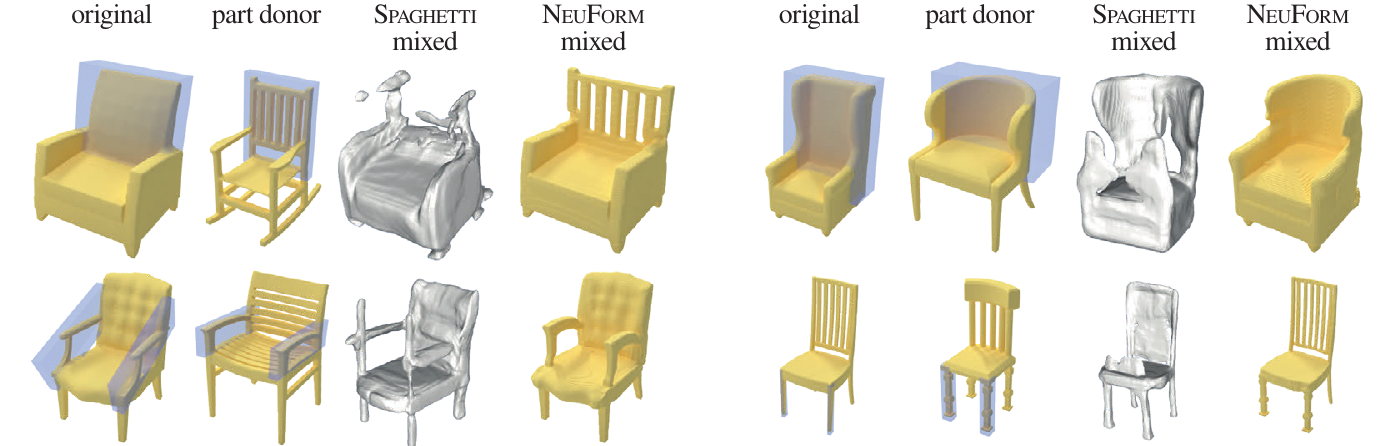}
    \caption{
    \textbf{Shape mixing.} 
    Mixing parts of different PartNet \texttt{chairs}. We replace the highlighted part in the original shape with the highlighted part in the donor shape, and compare our results to \spaghetti on  re-mixed shapes. Similar to the editing setting, \spaghetti's quality deteriorates on shapes with mixed parts. \name combines the foreign part more seamlessly into the shape.}
    \label{fig:mixing}
\end{figure}

Table~\ref{tab:comparison_chair_recons} shows quantitative results of this comparison and Figure~\ref{fig:comparison_chair_recons} shows qualitative examples for all methods. \spaghetti performs well in joint regions, but since it is a generalizable model, it lags behind the overfitted model and \coalesce in non-joint regions, giving a lower performance overall. \coalesce has the lowest performance in joint regions, as it struggles with larger or more extended joint areas, and has reasonable performance in non-joint areas. While \coalesce uses the ground truth geometry in non-joint areas, some of the joint geometry tends to incorrectly extend into the non-joint areas, lowering the performance.
As expected, our generalizable representation performs well in joint regions, and misses detail in non-joint regions. In this reconstruction task, the overfitted representation performs significantly better in joint regions than in the edit tasks we describe in the next sections, since the part configuration of the reconstructed shape is the same as the part configuration it was overfitted to. In the reconstructions, errors at the joints are due to the missing ground truth in joint regions. \name combines the advantages of the overfitted- and the generalizable representations, producing both plausible joints and detailed geometry. 

\paragraph{(ii) Shape Editing.}
\label{sec:shape_editing}

We experiment with shape edits by modifying the parameters of one or multiple cuboids of our shape representation. Editing results of \name compared to the generalizable and overfitted representations are shown in Figure~\ref{fig:comparison_chair_edits}. Edits on the generalizable representation confirm the trend we saw in the reconstruction task: joints are plausible after edits, but geometric detail is not preserved. When editing the overfitted representation, we observe significant artifacts near the joints, due to the previously unseen part configuration. Our adaptive overfitting strategy preserves the plausible joints of the generalizable representation as well as the geometric detail of the overfitted representation.

In Figure~\ref{fig:comparison_chair_edits_coalesce_spaghetti}, we compare shape editing to \coalesce and \spaghetti. Since \coalesce does not supports fine-grained edits, and \spaghetti does not support scaling, we compare to each on a separate set of edits. As we saw in the reconstruction, \coalesce struggles with extended joints, while \spaghetti's geometry deteriorates significantly after an edit. 

\paragraph{(iii) Shape Mixing.}
We demonstrate our model's ability to assemble new shapes from the parts of pre-existing ones in Figure~\ref{fig:mixing}. We mix and match cuboids and their associated part features from different \texttt{chairs}, and then blend the parts together. For a given query point and its closest part $P$, we use the overfitted representation associated with the shape that $P$ was originally part of. Our method synthesizes much smoother joint connections between parts while preserving their surface details.

\label{sec:shape_part_mixing}

\paragraph{Additional shape categories.}

\begin{figure}[t]
    \centering
    \includegraphics[width=\linewidth]{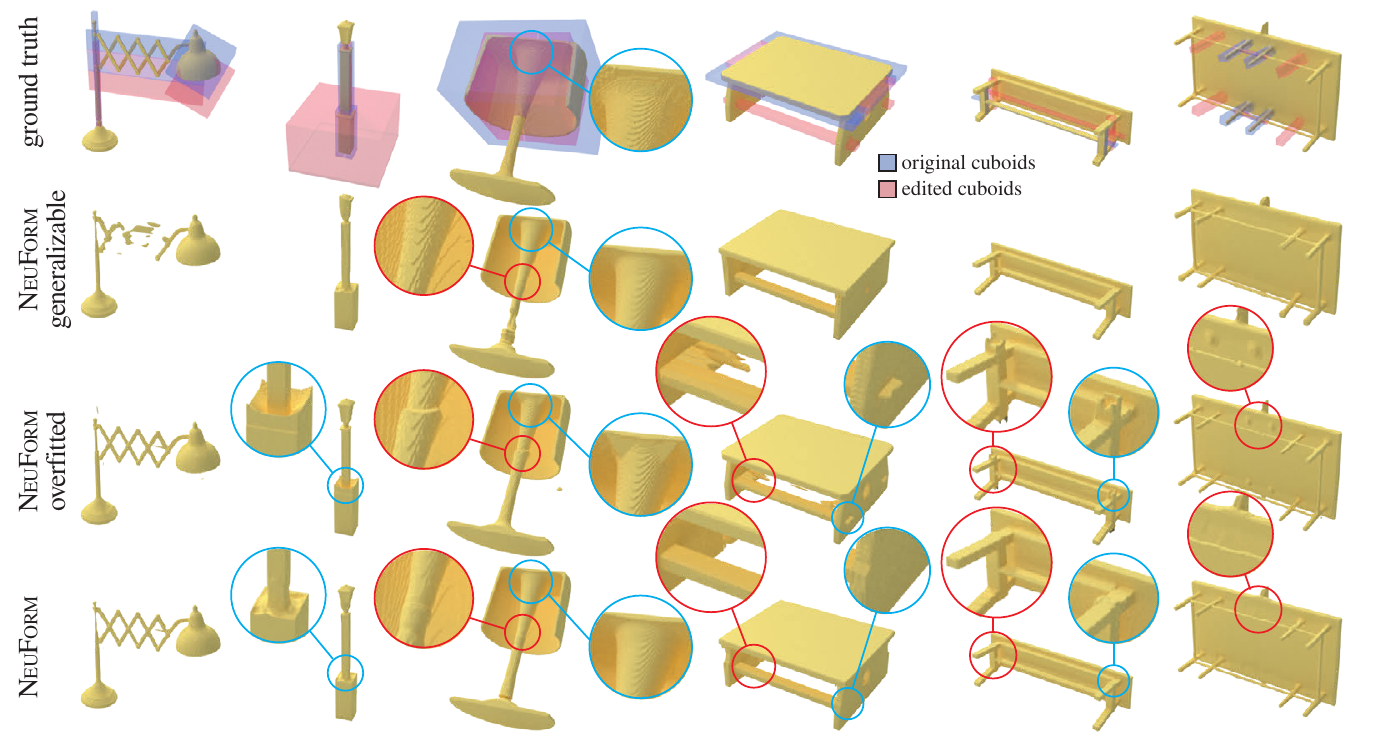}
    \caption{
    \textbf{Different object categories.} 
    Shape edits on PartNet \texttt{tables} and \texttt{lamps}. 
    Similar to \texttt{chairs}, the generalizable model lacks detail and the overfitted model contains artifacts in joint regions, whereas \name combines the advantages of both.}
    \label{fig:other_cat_edits}
\end{figure}



Figure~\ref{fig:other_cat_edits} shows edit results on \texttt{tables} and \texttt{lamps}, compared to the generalizable and overfitted representations. Similar to \texttt{chairs}, the generalizable representation is missing shape detail, resulting, for example, in artifacts on thin parts, while the overfitted representation struggles with joint areas. In the right-most \texttt{table}, we can clearly see that these artifacts occur both in regions that are joints after the edit, as well as regions that used to be joints in the original shape. Adaptive overfitting avoids these artifacts.

\section{Conclusions}

We have introduced the \name architecture to enable adaptive mixing of information between a generalizable neural neural network, trained on a collection of shapes, and an overfitted model, trained on a single shape to capture its idiosyncrasies. We achieved this by designing a network architecture that allows adaptive mixing of networks by carefully blending respective network weights and training history.

Our work is just the first step in the direction of merging overfitted and generalizable models. For example, currently the two models do not have explicit knowledge of each other, adding this knowledge could be interesting future work. For shape editing, this could allow the generalizable network to focus more on joint geometry. Another limitation is the currently non-data-driven blending field. Learning a context-based blending factor is a promising next step for facilitating easier and higher quality editing.

% ...

{
\small
\bibliographystyle{acm}
\bibliography{main}

\begin{thebibliography}{10}

\bibitem{achlioptas2018latent_pc}
{\sc Achlioptas, P., Diamanti, O., Mitliagkas, I., and Guibas, L.~J.}
\newblock Learning representations and generative models for 3d point clouds.
\newblock {\em ICML\/} (2018).

\bibitem{atzmon_sal_2019}
{\sc Atzmon, M., and Lipman, Y.}
\newblock {SAL}: Sign agnostic learning of shapes from raw data.
\newblock In {\em Proceedings of the IEEE/CVF Conference on Computer Vision and
  Pattern Recognition\/} (2020), pp.~2565--2574.

\bibitem{atzmon_sal_2020}
{\sc Atzmon, M., and Lipman, Y.}
\newblock {SAL}++: Sign agnostic learning with derivatives.
\newblock {\em arXiv preprint arXiv:2006.05400\/} (2020).

\bibitem{bednarik_shape_2019}
{\sc Bednarik, J., Parashar, S., Gundogdu, E., Salzmann, M., and Fua, P.}
\newblock Shape reconstruction by learning differentiable surface
  representations.
\newblock In {\em Proceedings of the IEEE/CVF Conference on Computer Vision and
  Pattern Recognition\/} (2020), pp.~4716--4725.

\bibitem{BrockLRW16}
{\sc Brock, A., Lim, T., Ritchie, J.~M., and Weston, N.}
\newblock Generative and discriminative voxel modeling with convolutional
  neural networks.
\newblock {\em CoRR\/} (2016).

\bibitem{chen_cvpr19}
{\sc Chen, Z., and Zhang, H.}
\newblock Learning implicit fields for generative shape modeling.
\newblock In {\em IEEE Computer Vision and Pattern Recognition (CVPR)\/}
  (2019).

\bibitem{dai2019scan2mesh}
{\sc Dai, A., and Nie{\ss}ner, M.}
\newblock Scan2mesh: From unstructured range scans to 3d meshes.
\newblock In {\em Proc. Computer Vision and Pattern Recognition (CVPR), IEEE\/}
  (2019).

\bibitem{dai2017complete}
{\sc Dai, A., Qi, C.~R., and Nie{\ss}ner, M.}
\newblock Shape completion using 3d-encoder-predictor cnns and shape synthesis.
\newblock {\em Proc. Computer Vision and Pattern Recognition (CVPR), IEEE\/}
  (2017).

\bibitem{davies2020overfit}
{\sc Davies, T., Nowrouzezahrai, D., and Jacobson, A.}
\newblock Overfit neural networks as a compact shape representation, 2020.

\bibitem{deprelle_learning_2019}
{\sc Deprelle, T., Groueix, T., Fisher, M., Kim, V.~G., Russell, B.~C., and
  Aubry, M.}
\newblock Learning elementary structures for 3d shape generation and matching.
\newblock {\em arXiv preprint arXiv:1908.04725\/} (2019).

\bibitem{genova2019}
{\sc Genova, K., Cole, F., Vlasic, D., Sarna, A., Freeman, W.~T., and
  Funkhouser, T.}
\newblock Learning shape templates with structured implicit functions.
\newblock In {\em ICCV\/} (2019).

\bibitem{GirdharFRG16}
{\sc Girdhar, R., Fouhey, D.~F., Rodriguez, M., and Gupta, A.}
\newblock Learning a predictable and generative vector representation for
  objects.
\newblock {\em CoRR abs/1603.08637\/} (2016).

\bibitem{gropp_implicit_2020}
{\sc Gropp, A., Yariv, L., Haim, N., Atzmon, M., and Lipman, Y.}
\newblock Implicit geometric regularization for learning shapes.
\newblock {\em arXiv preprint arXiv:2002.10099\/} (2020).

\bibitem{groueix2018papier}
{\sc Groueix, T., Fisher, M., Kim, V.~G., Russell, B.~C., and Aubry, M.}
\newblock A papier-m{\^a}ch{\'e} approach to learning 3d surface generation.
\newblock In {\em Proceedings of the IEEE conference on computer vision and
  pattern recognition\/} (2018), pp.~216--224.

\bibitem{dualSDF20}
{\sc Hao, Z., Averbuch-Elor, H., Snavely, N., and Belongie, S.}
\newblock Dualsdf: Semantic shape manipulation using a two-level
  representation, 2020.

\bibitem{hertz2022spaghetti}
{\sc Hertz, A., Perel, O., Giryes, R., Sorkine-Hornung, O., and Cohen-Or, D.}
\newblock Spaghetti: Editing implicit shapes through part aware generation.
\newblock {\em arXiv preprint arXiv:2201.13168\/} (2022).

\bibitem{huang2018watertight}
{\sc Huang, J., Su, H., and Guibas, L.}
\newblock Robust watertight manifold surface generation method for shapenet
  models.
\newblock {\em arXiv preprint arXiv:1802.01698\/} (2018).

\bibitem{kingma2015adam}
{\sc Kingma, D.~P., and Ba, J.}
\newblock Adam: A method for stochastic optimization.
\newblock In {\em ICLR (Poster)\/} (2015).

\bibitem{littwin_deep_2019}
{\sc Littwin, G., and Wolf, L.}
\newblock Deep meta functionals for shape representation.
\newblock In {\em Proceedings of the IEEE/CVF International Conference on
  Computer Vision\/} (2019), pp.~1824--1833.

\bibitem{liu2018voxelgan}
{\sc Liu, J., Yu, F., and Funkhouser, T.}
\newblock Interactive 3d modeling with a generative adversarial network.
\newblock {\em International Conference on 3D Vision (3DV)\/} (2017).

\bibitem{martel2021acorn}
{\sc Martel, J.~N., Lindell, D.~B., Lin, C.~Z., Chan, E.~R., Monteiro, M., and
  Wetzstein, G.}
\newblock Acorn: Adaptive coordinate networks for neural scene representation.
\newblock {\em arXiv preprint arXiv:2105.02788\/} (2021).

\bibitem{occupancy_nets_2019}
{\sc Mescheder, L., Oechsle, M., Niemeyer, M., Nowozin, S., and Geiger, A.}
\newblock Occupancy networks: Learning 3d reconstruction in function space.
\newblock In {\em Proceedings IEEE Conf. on Computer Vision and Pattern
  Recognition (CVPR)\/} (2019).

\bibitem{mildenhall_nerf_2020}
{\sc Mildenhall, B., Srinivasan, P.~P., Tancik, M., Barron, J.~T., Ramamoorthi,
  R., and Ng, R.}
\newblock {NeRF}: Representing scenes as neural radiance fields for view
  synthesis.
\newblock In {\em European Conference on Computer Vision\/} (2020), Springer,
  pp.~405--421.

\bibitem{MoGuerreroEtAl:StructEdit:Arxiv:2019}
{\sc Mo, K., Guerrero, P., Yi, L., Su, H., Wonka, P., Mitra, N., and Guibas,
  L.}
\newblock Structedit: Learning structural shape variations.
\newblock {\em arXiv preprint arXiv:1908.00575\/} (2019).

\bibitem{MoGuerreroEtAl:StructureNet:SiggraphAsia:2019}
{\sc Mo, K., Guerrero, P., Yi, L., Su, H., Wonka, P., Mitra, N., and Guibas,
  L.}
\newblock Structurenet: Hierarchical graph networks for 3d shape generation.
\newblock {\em ACM TOG\/} (2019).

\bibitem{Mo_2019_CVPR_partnet}
{\sc Mo, K., Zhu, S., Chang, A.~X., Yi, L., Tripathi, S., Guibas, L.~J., and
  Su, H.}
\newblock {PartNet}: A large-scale benchmark for fine-grained and hierarchical
  part-level {3D} object understanding.
\newblock In {\em The IEEE Conference on Computer Vision and Pattern
  Recognition (CVPR)\/} (June 2019).

\bibitem{morreale2022neuralConvolutionalSurfaces}
{\sc Morreale, L., Aigerman, N., Guerrero, P., Kim, V.~G., and Mitra, N.~J.}
\newblock Neural convolutional surfaces.
\newblock In {\em Proc. {CVPR}\/} (2022).

\bibitem{morreale2021neural}
{\sc Morreale, L., Aigerman, N., Kim, V.~G., and Mitra, N.~J.}
\newblock Neural surface maps.
\newblock In {\em Proceedings of the IEEE/CVF Conference on Computer Vision and
  Pattern Recognition\/} (2021), pp.~4639--4648.

\bibitem{park2019deepsdf}
{\sc Park, J.~J., Florence, P., Straub, J., Newcombe, R., and Lovegrove, S.}
\newblock Deepsdf: Learning continuous signed distance functions for shape
  representation.
\newblock In {\em Proceedings of the IEEE/CVF Conference on Computer Vision and
  Pattern Recognition\/} (2019), pp.~165--174.

\bibitem{Poursaeed20a}
{\sc Poursaeed, O., Fisher, M., Aigerman, N., and Kim, V.~G.}
\newblock Coupling explicit and implicit surface representations for generative
  3d modeling.
\newblock {\em ECCV\/} (2020).

\bibitem{pointnet}
{\sc Qi, C.~R., Su, H., Mo, K., and Guibas, L.~J.}
\newblock Pointnet: Deep learning on point sets for 3d classification and
  segmentation, 2016.

\bibitem{Sinha2016DeepL3}
{\sc Sinha, A., Bai, J., and Ramani, K.}
\newblock Deep learning 3d shape surfaces using geometry images.
\newblock In {\em ECCV\/} (2016).

\bibitem{sitzmann_implicit_2020}
{\sc Sitzmann, V., Martel, J.~N., Bergman, A.~W., Lindell, D.~B., and
  Wetzstein, G.}
\newblock Implicit neural representations with periodic activation functions.
\newblock {\em arXiv preprint arXiv:2006.09661\/} (2020).

\bibitem{Su2017PointGen}
{\sc Su, H., Fan, H., and Guibas, L.}
\newblock A point set generation network for 3d object reconstruction from a
  single image.
\newblock {\em CVPR\/} (2017).

\bibitem{deformSyncNet:2020}
{\sc Sung, M., Jiang, Z., Achlioptas, P., Mitra, N.~J., and Guibas, L.~J.}
\newblock Deformsyncnet: Deformation transfer via synchronized shape
  deformation spaces, 2020.

\bibitem{takikawa2021neural}
{\sc Takikawa, T., Litalien, J., Yin, K., Kreis, K., Loop, C., Nowrouzezahrai,
  D., Jacobson, A., McGuire, M., and Fidler, S.}
\newblock Neural geometric level of detail: Real-time rendering with implicit
  3d shapes.
\newblock In {\em Proc. {CVPR}\/} (2021), pp.~11358--11367.

\bibitem{trimesh}
{\sc trimesh}.
\newblock Trimesh [https://trimsh.org/], 2022.

\bibitem{vaswani_attention_2017}
{\sc Vaswani, A., Shazeer, N., Parmar, N., Uszkoreit, J., Jones, L., Gomez,
  A.~N., Kaiser, L., and Polosukhin, I.}
\newblock Attention is all you need.
\newblock {\em arXiv preprint arXiv:1706.03762\/} (2017).

\bibitem{yang_foldingnet_2018}
{\sc Yang, Y., Feng, C., Shen, Y., and Tian, D.}
\newblock {FoldingNet}: Point cloud auto-encoder via deep grid deformation.
\newblock In {\em Proceedings of the IEEE Conference on Computer Vision and
  Pattern Recognition\/} (2018), pp.~206--215.

\bibitem{yifan2021geometryconsistent}
{\sc Yifan, W., Rahmann, L., and Sorkine-Hornung, O.}
\newblock Geometry-consistent neural shape representation with implicit
  displacement fields, 2021.

\bibitem{yin2020coalesce}
{\sc Yin, K., Chen, Z., Chaudhuri, S., Fisher, M., Kim, V.~G., and Zhang, H.}
\newblock Coalesce: Component assembly by learning to synthesize connections.
\newblock In {\em 2020 International Conference on 3D Vision (3DV)\/} (2020),
  IEEE, pp.~61--70.

\end{thebibliography}
}






\appendix

\section{Overview}
In this supplementary document, we provide additional details on our data preparation procedure (Section~\ref{sec:data_preparation}), our architecture (Section~\ref{sec:architecture_details}), and the baselines (Section ~\ref{sec:baseline_details}). Additionally, we provide a quantitative evaluation of the shape editing experiments (Section~\ref{sec:shape_editing_quantitative}), extend the shape mixing experiments to include a comparison to \textsc{Coalesce} (Section~\ref{sec:part_mixing_coalesce}), and provide several additional ablations of our approach (Section~\ref{sec:ablations}).

\section{Data Preparation}
\label{sec:data_preparation}

We use the PartNet~\cite{Mo_2019_CVPR_partnet} dataset for training and evaluation. The PartNet dataset defines a hierarchical decomposition of each shape into parts. We use the first level of the part hierarchy except for the base of chairs and tables, where we use a deeper level to obtain individual legs and bars. To compute ground truth occupancy, we make the shapes watertight using an existing method~\cite{huang2018watertight}. Next, we center each shape at the origin and scale it such that the largest extent along any axis is $[-1, 1]$. We fit oriented boundary boxes to parts using TriMesh~\cite{trimesh}. To obtain a point cloud for the part geometry encoder $h_{\psi}$, we uniformly sample 5k surface and volume points for each part, additionally storing the SDF gradient for each point (the SDF gradient generalizes the surface normal to the volume), and transforming the resulting point cloud into the local coordinate frame of the part's cuboid. Finally, to obtain the query points $x$ used during training, we sample 50k points uniformly inside the bounding cube $[-1, 1]^3$ and 50k points on the surface of the mesh with added Gaussian noise of $\sigma = 0.05$.

\section{Architecture Details}
\label{sec:architecture_details}

\paragraph{Part Encoder $h$.}
We encode the geometry of each part using a PointNet~\cite{pointnet} encoder consisting of six hidden layers prior to the Max-Pooling layer and two hidden layers after pooling. The hidden layers start with dimensionality of 64 and consecutively double until reaching dimensionality 512. As input, we randomly sample a subset of 4096 surface and volume points for each part, taken from the point cloud we pre-computed during the data preparation step (see Section~\ref{sec:data_preparation}). We input both the point locations and SDF gradients, resulting in a 6-dimensional input vector per point. The output is a $512$-dimensional feature vector $g_i$ per part that captures the part geometry in the local coordinate frame of the part's cuboid.

\paragraph{Part mixing network $f^m$.}
The part mixing network performs two main operations: it first combines the geometry feature vector $g_i$ and the cuboid parameters $C_i$ of each part into a per-part feature vector $p'_i$, and then exchanges information between parts using a self-attention layer to obtain an updated per-part feature vector $p^\mathcal{P}_i$. To obtain the per-part feature vector $p'_i$, a $512$-dimensional cuboid feature vector $c_i$ is computed from the cuboid parameters $C_i$ using a three-layer MLP with $512$ hidden dimensions, and then added to the feature vector $g_i$: $p'_i = g_i + c_i$. Similar to SPAGHETTI~\cite{hertz2022spaghetti}, we use multiple self-attention layers to mix information between the per-part feature vectors $p'_i$ to obtain updated feature vectors $p^\mathcal{P}_i$:
\begin{equation}
    \{p^\mathcal{P}_i\}_i = \mathtt{SAtt}^4(\{p'_i\}_i) 
\end{equation}
where $\mathtt{SAtt}^4$ denotes four Transformer~\cite{vaswani_attention_2017} self-attention blocks.
Each block includes an attention layer with 8 attention heads, followed by a feed-forward layer. See the original Transformers paper~\cite{vaswani_attention_2017} for details.


\paragraph{Part query network $f^x$.} The part query network queries all parts at a query point $x$ by performing cross-attention from the query point to all parts. Since the part geometry is defined in local coordinates of the part's cuboid, we transform the query point into local coordinates $x^l_i = T^{-1}_{C_i}(x)$, where $T^{-1}_{C_i}$ is the transformation to the local coordinate frame of the cuboid $C_i$. We use a learned positional encoding $\pi$ for the local coordinates $x^l_i$, and perform cross-attention from each encoded local coordinate to all cuboids:
\begin{equation}
    \{q^\mathcal{P}_i(x)\} = \mathtt{CAtt}^4(\{\pi(x^l_i)\}, \{p^\mathcal{P}_1 + b_0, \dots, p^\mathcal{P}_i + b_1, \dots, p^\mathcal{P}_n + b_0\}),
\end{equation}
where $\mathtt{CAtt}^4(a,b)$ denotes four Transformer cross-attention blocks, with queries based on $a$ and keys/values based on $b$. Each block includes an attention layer with 8 attention heads followed by a feed-forward layer. See the original Transformers paper~\cite{vaswani_attention_2017} for details. Note that a different set of keys/values is used for each query, since the indicator feature $b_1$ is added to a different part feature vector for each local query point: for query point $x^l_i$, it is added to the part feature vector $p^\mathcal{P}_i$.

\paragraph{Global occupancy network $f^o$.} The global occupancy network is implemented as a two-layer MLP with 512 hidden dimensions.




\section{Baseline Details}
\label{sec:baseline_details}
\paragraph{\textsc{Coalesce}~\cite{yin2020coalesce}.} We use the pre-trained model provided by the authors and pre-process all shapes using the approach described in \textsc{Coalesce}, making sure to re-normalize the shapes so the scaling and orientation is comparable to the existing test set shapes. Since our cuboids use a more fine-grained shape decomposition than \textsc{Coalesce}, we assign each of our cuboids to one of the shape parts defined by \textsc{Coalesce} and treat each resulting group of cuboids as a single part. We then define a segmentation of the shape by assigning each surface point to the cuboid it has the smallest signed distance to and remove the surface within a small radius of segment boundaries, as described in the \textsc{Coalesce} paper. The output of \textsc{Coalesce} is transformed back to our normalized coordinates for comparison with the ground truth.

\paragraph{\textsc{Spaghetti}~\cite{hertz2022spaghetti}.} Here, we also use the pre-trained model provided by the authors and make sure to normalize the shapes as required by \textsc{Spaghetti}. Unlike \textsc{Coalesce}, we can work directly with our cuboids, as \textsc{Spaghetti} can handle fine-grained parts and shares our cuboid representation. When editing or mixing shapes, we use the editing UI provided by the authors (we do not need to use the UI for shape reconstruction). The output of \textsc{Spaghetti} is transformed back to our normalized coordinates for comparison with the ground truth.

\section{Quantitative Evaluation of Shape Edits}
\label{sec:shape_editing_quantitative}

\begin{table}[tb]
  \small
  \centering
  \caption{\textbf{Quantitative comparison of chair edits.} We show a quantitative evaluation of the edits shown in Figure~\ref{fig:comparison_chair_edits} of the main paper. The generalizable model performs better than the overfitted model in joint regions, while the reverse is true for non-joint regions.  \name combines the advantages of both an performs better on average in all regions.}
  \setlength{\tabcolsep}{2pt}
  \begin{tabular}{r @{\hskip 20pt} ccc @{\hskip 20pt} ccc @{\hskip 20pt} ccc}
    \toprule
    & \multicolumn{3}{c}{Edited joint regions} & \multicolumn{3}{c}{Non-joint regions} & \multicolumn{3}{c}{All regions}\\
    \midrule
    & CD$\downarrow$ & EMD$\downarrow$ & SDF$\downarrow$ & CD$\downarrow$ & EMD$\downarrow$ & SDF$\downarrow$ &  CD$\downarrow$ & EMD$\downarrow$ & SDF$\downarrow$ \\ 
    \midrule
    \ourname generalizable & 0.052 & \textbf{48.07} & 0.745 & 0.042 & 72.31 & 1.751 & 0.047 & 60.19 & 1.248 \\
    \ourname overfitted & 0.110 & 64.89 & 1.159 & 0.020 & 69.12 & 0.687 & 0.065 & 67.00 & 0.923 \\
    \ourname & \textbf{0.052} & 49.72 & \textbf{0.642} & \textbf{0.019} & \textbf{62.66} & \textbf{0.684} & \textbf{0.036} & \textbf{56.19} & \textbf{0.663} \\
    \bottomrule
  \end{tabular}
  \label{tab:comparison_chairs_overfit_recon} 
\end{table}

In this section, we show a quantitative evaluation of the edits shown in Figure~\ref{fig:comparison_chair_edits} of the main paper. Since we do not have ground truth for a shape with an edited cuboid configuration, we do the inverse: we start with the edited cuboid configuration and overfit to it (i.e. ($\frozen{\partset}, \partset) = (\edited{\partset}, \original{\partset})$ instead of $(\frozen{\partset}, \partset) = (\original{\partset}, \edited{\partset})$ as described at the end of Section~\ref{sec:adaptive} in the main paper). Then, we re-arrange the edited cuboids to undo the edit. Since this should result in the original shape, we do have ground truth for this re-arrangement that we can use to compute the quantitative metrics defined in Section~\ref{sec:results} of the main paper. Note that it is possible to overfit to the edited cuboid configuration, since our overfitted model only requires ground truth in non-joint regions for training, which we can obtain by simply transforming individual parts geometries.

Results are shown in Table~\ref{tab:comparison_chairs_overfit_recon}. We can see that the generalizable model performs better than the overfitted model in joint regions, while the reverse is true for non-joint regions. \name combines the advantages of both an performs better on average in all regions. Note that \name even slightly outperforms the generalizable model in joint regions and the overfitted model in non-joint regions, since both the joint regions and the non-joint regions include small transition regions between joints and non-joints that \name performs better on than either overfitted or generalizable model alone.

\section{Part Mixing Comparison to COALESCE}
\label{sec:part_mixing_coalesce}

\begin{figure}[t]
    \centering
    \includegraphics[width=\linewidth]{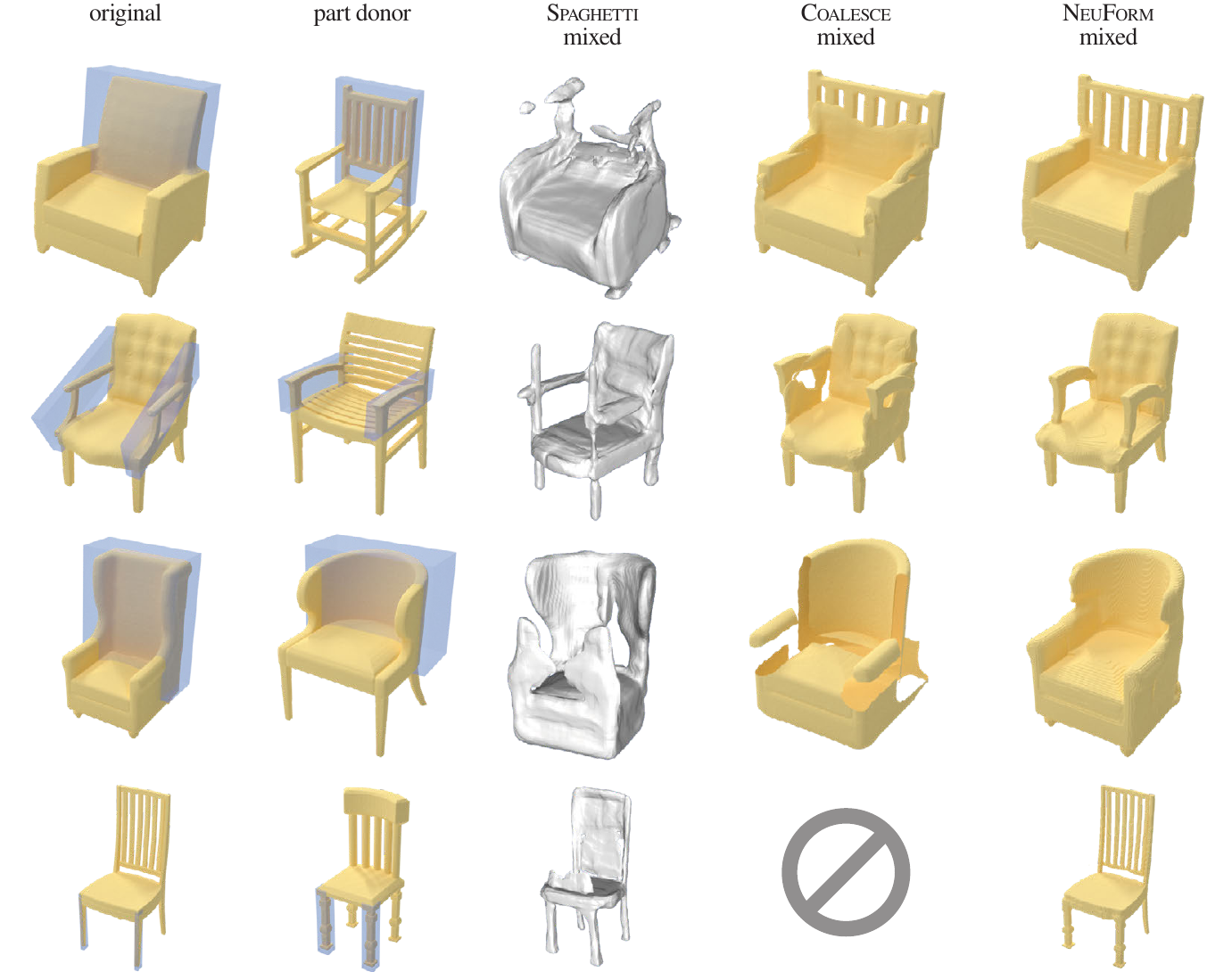}
    \caption{
    \textbf{Shape mixing with \textsc{Coalesce}.} 
    We extend the shape mixing results shown in Figure~\ref{fig:mixing} of the main paper by adding a comparison to \textsc{Coalesce}. Joints produced by \textsc{Coalesce} are generally more noisy. In row three, one of the steps in the pipeline of \textsc{Coalesce} fails, producing no joint geometry and the edit in row four \textsc{Coalesce} is not applicable, since fine-grained part edits like individual chair legs are not supported. \name produces more plausible results with fewer artifacts.}
    \label{fig:comparison_chair_mixing_with_coalesce}
\end{figure}

In Figure~\ref{fig:comparison_chair_mixing_with_coalesce}, we extend the part mixing experiments shown in the main paper with a comparison to \textsc{Coalesce}~\cite{yin2020coalesce}. Similar to the editing results in Figure~\ref{fig:comparison_chair_edits_coalesce_spaghetti} of the main paper, we can see that \textsc{Coalesce} preserves geometric detail of individual parts. But as the \textsc{Coalesce} authors note in their limitations, the method struggles to connect parts with stronger geometric or topological incompatibility, resulting in noisy joints four our shape mixing examples. In the example in row three, the Poisson blending step of \textsc{Coalesce} fails, completely removing any joint geometry. In row four, we can see another limitation of \textsc{Coalesce} that the authors point out in their paper: editing or mixing fine-grained parts like individual chair legs is not supported, due to the larger inconsistency between the part decompositions of different chairs in the dataset when using more fine-grained parts.

\section{Additional Ablations}
\label{sec:ablations}

\begin{figure}[t]
    \centering
    \includegraphics[width=\linewidth]{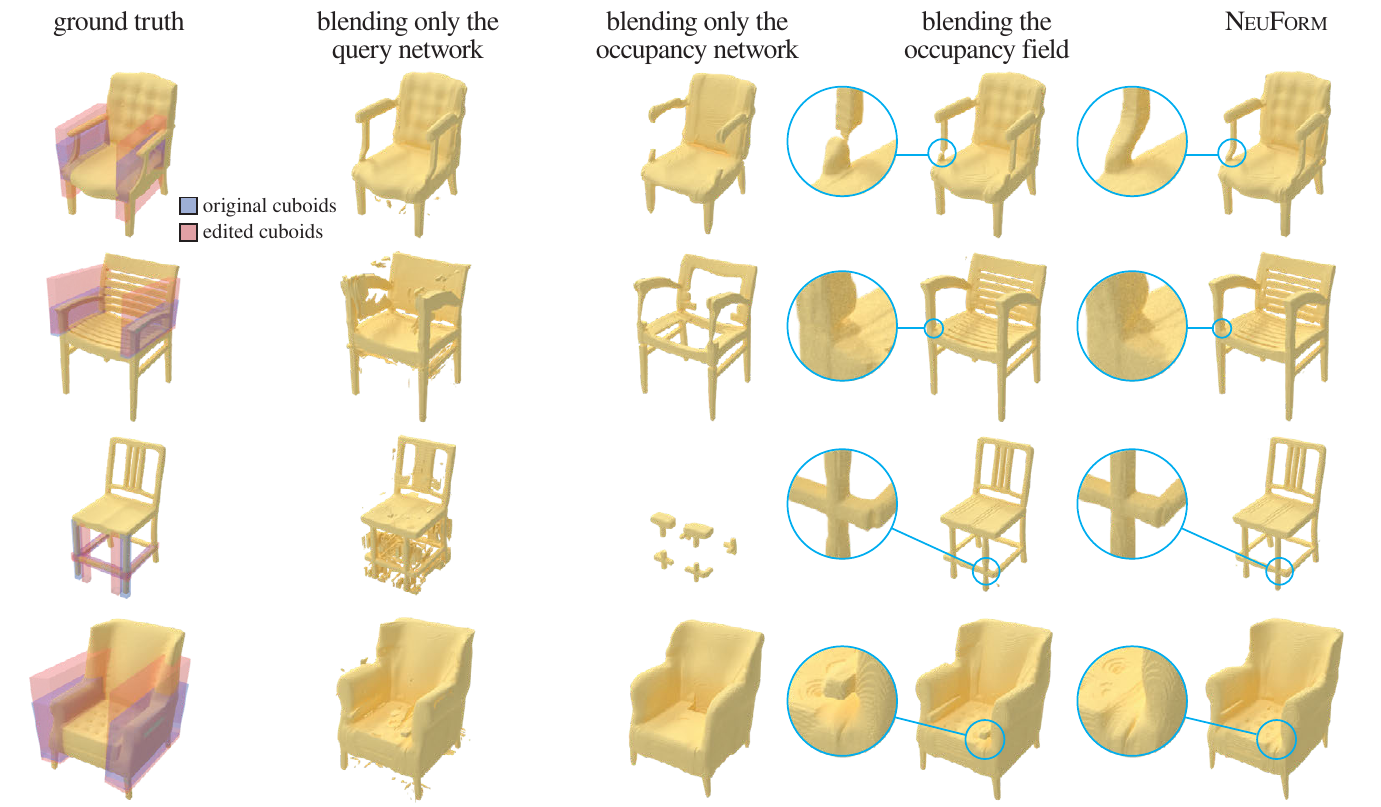}
    \caption{
    \textbf{Additional ablations.} We show ablations that only blend with the overfitted network for a subset of the networks (second column: only the part query network $f^x$, third column: only the global occupancy network $f^o$), and an ablation that directly blends the occupancy fields (fourth column). Blending only a subset of networks results in severe artifacts, while directly blending the occupancy fields gives overall better results, but shows artifacts in joints where there is larger disagreement between the generalizable and the overfitted representations.}
    \label{fig:ablations}
\end{figure}

We show three additional ablations qualitatively in Figure~\ref{fig:ablations}. First we show the effect of only blending a subset of our networks instead of blending both $f^x$ and $f^o$. Results are shown in the second and third columns. Since this results in a parameter combinations that were not seen during training, results show severe artifacts.
Next, we show a possible alternative to blending the overfitted and generalizable representations in network parameter space: we show directly blending the occupancy fields output by the two representations. This seems to work well at first glance, but on closer inspection, we can see that it results in artifacts in regions with larger disagreement between the overfitted and the generalizable representations. For example, this is clearly visible in the region highlighted in Figure~\ref{fig:ablations}, first row, fourth column. The results of \name show that blending in the network parameter space handles disagreement between the overfitted and the generalizable representations more gracefully.


\end{document}